\documentclass[10pt,twocolumn,letterpaper]{article}

\usepackage[pagenumbers]{cvpr} % To force page numbers, e.g. for an arXiv version

\usepackage[dvipsnames]{xcolor}

\newcommand{\sh}[1]{\textcolor{black}{#1}}
\newcommand{\nj}[1]{\textcolor{black}{#1}}
\usepackage{kotex}

\definecolor{cvprblue}{rgb}{0.21,0.49,0.74}
\usepackage[pagebackref,breaklinks,colorlinks,citecolor=cvprblue]{hyperref}
\usepackage{multirow}
\usepackage{multicol}
\usepackage{comment}
\usepackage{makecell}
\usepackage{hhline}

\def\paperID{19} % *** Enter the Paper ID here
\def\confName{CVPR}
\def\confYear{2024}

\title{Unleash the Potential of CLIP for Video Highlight Detection}

\author{Donghoon Han$^{1*}$ \quad
Seunghyeon Seo$^{2*}$ \quad
Eunhwan Park$^{1}$ \quad
Seong-Uk Nam$^1$ \quad
Nojun Kwak$^{2\dagger}$
\smallskip
\\
$^1$Buzzni AI Lab \quad
$^2$Seoul National University \\
{\tt\small \{owen, jude, zaid\}@buzzni.com} \quad
{\tt\small \{zzzlssh, nojunk\}@snu.ac.kr}
}

\begin{document}
\maketitle
{\let\thefootnote\relax\footnotetext{{$^{*}$Equal contribution.
$^{\dagger}$Corresponding author.}}}
\begin{abstract}
%As multimodal and large language models (LLMs) harness open-world knowledge, these models opened up new possibilities for various tasks and applications. The video domain is one of the domains that gained the most from those two. In this paper, we introduce Highlight-CLIP (HL-CLIP) to perform the video highlight detection task by fully harnessing the pre-trained knowledge of multimodal models. By simply finetuning the multimodal encoder and our saliency pooling technique, we have achieved so far the best performance on the highlight detection task, QVHighlight Benchmark, to the authors' knowledge.
\nj{Multimodal and large language models (LLMs) have revolutionized the utilization of open-world knowledge, unlocking novel potentials across various tasks and applications. Among these domains, the video domain has notably benefited from their capabilities. In this paper, we present Highlight-CLIP (HL-CLIP), a method designed to excel in the video highlight detection task by leveraging the pre-trained knowledge embedded in multimodal models. By simply fine-tuning the multimodal encoder \nj{in combination with} our innovative saliency pooling technique, we have achieved the state-of-the-art performance in the highlight detection task, the QVHighlight Benchmark, to the best of our knowledge.}
\end{abstract}    
\section{Introduction}
\label{sec:intro}
\begin{comment}
    https://arxiv.org/abs/2208.03550
    https://arxiv.org/abs/2103.00020
    https://openaccess.thecvf.com/content/WACV2024/papers/Luo_Zero-Shot_Video_Moment_Retrieval_From_Frozen_Vision-Language_Models_WACV_2024_paper.pdf
    https://openaccess.thecvf.com/content/ICCV2023W/MMFM/papers/Pan_Retrieving-to-Answer_Zero-Shot_Video_Question_Answering_with_Frozen_Large_Language_Models_ICCVW_2023_paper.pdf
    
    % Large Scale Pre-training 이 왜 도움이 되는지 분석되는 논문들 인용    https://openaccess.thecvf.com/content/CVPR2023/papers/Weers_Masked_Autoencoding_Does_Not_Help_Natural_Language_Supervision_at_Scale_CVPR_2023_paper.pdf

\end{comment}

% such as 부분 인용 추가하는 게 좋음.
\sh{
Recent advancements in the Natural Language Processing (NLP) area have been significantly revolutionized by the adoption of the pre-training method on large-scale text corpus. Based on the results that have convincingly demonstrated \nj{their} superiority of training on large-scale datasets, recent pioneering works on multimodal models have unveiled significant capabilities for zero-shot text-image matching. Thanks to the zero-shot text-image matching ability, multimodal models have shown \nj{promising results across various} downstream tasks, such as visual question answering~\cite{tiong2022plug}, image captioning~\cite{DBLP:conf/iclr/LiZW023}, text-image retrieval~\cite{Baldrati_2023_ICCV}, etc.}

\sh{
Despite these improvements, the pre-training method mainly relies on ``image–text matching'', meaning that the multimodal models have often suffered from the lack of spatial and temporal knowledge. For example, tasks such as detecting highlights in videos not only require recognizing objects and their descriptions but also understanding how these elements interact over \nj{time}. Our underlying assumption is \nj{that} integrating both temporal and spatial knowledge would enhance their performance on \nj{tasks that need temporal awareness}.}

\sh{
To this end, we introduce a finetuning strategy for video highlight detection task\nj{s}, namely Highlight-CLIP~(HL-CLIP). Different from prior studies, our \nj{aim} is exploiting the full potential of multimodal models.} Different from prior studies~\cite{lei2021detecting, liu2022umt, lin2023univtg, moon2023query, moon2023correlation}, we only utilize a pre-trained multimodal encoder to achieve \nj{better} performance in the video highlight detection task on the QVHighlight Benchmark~\cite{lei2021detecting}, thereby emphasizing the importance of utilizing the capability of pre-trained multimodal models.

\sh{The contributions of our paper can be briefly summarized as follows:
\begin{itemize}
    % \item We propose a simple finetuning strategy of CLIP for video highlight detection task.
    % \item We enhance the performance of CLIP-only framework for video highlight detection, coined as HL-CLIP, by a simple finetuning strategy.
    \item We propose HL-CLIP, a simple finetuning framework for video highlight detection task\nj{s}, which leads to enhanced training efficiency.
    \item We further boost the highlight detection performance at inference time without additional training by our proposed saliency pooling technique.
    \item Our HL-CLIP achieves the state-of-the-art performance on \nj{the QVHighlight benchmark, a} highlight detection task.
\end{itemize}}

\section{Related Works}
\label{sec:relworks}
\subsection{QVHighlight Dataset}
QVH\sh{ighlight} dataset~\cite{lei2021detecting} consists of over 10,000 videos with annotations, including human-written queries and corresponding segments marked with saliency ratings. The video samples range from lifestyle vlogs to news videos from YouTube, aiming to secure a wide range of content suitable for user annotation. Each video segment spans 150 seconds and features human-annotated free-form queries selected using targeted search queries such as `daily vlog' and `news hurricane.' This annotation process, conducted via Amazon Mechanical Turk, ensures robust data quality characterized by high inter-annotator agreement. Additionally, we assessed the saliency of selected video segments using a Likert-scale rating system, ensuring comprehensive evaluation by involving multiple workers for each clip.

\begin{figure*}[t!]
\centering
\includegraphics[width=.7\textwidth]{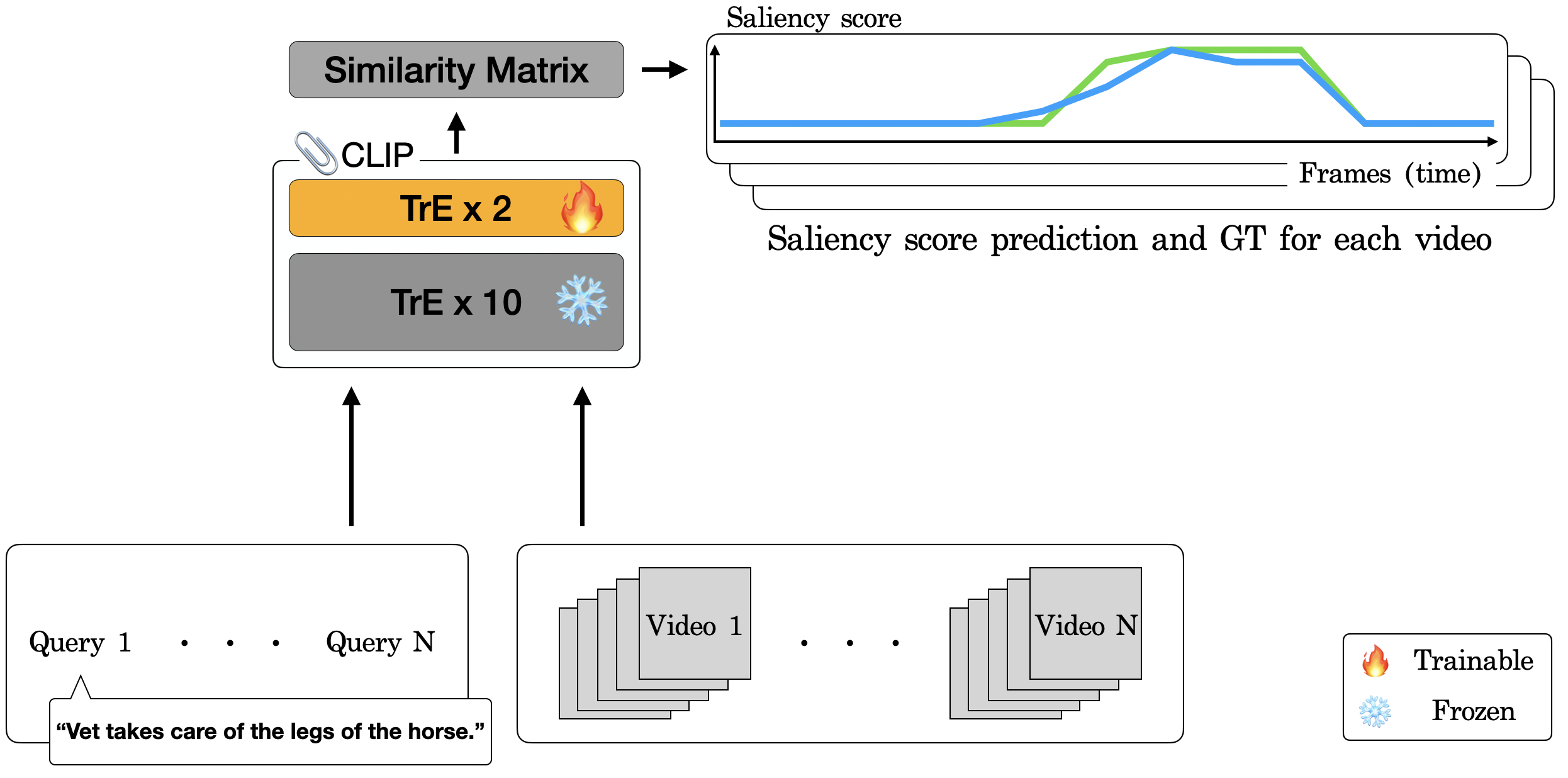}
\caption{The overall architecture of Highlight-CLIP~(HL-CLIP). \nj{The last} few layers of transformer encoders are finetuned to estimate \nj{the} saliency score between \nj{a} frame and \nj{a} query. \nj{The green and blue lines denote the predicted and the ground truth saliency score of a video frame given a query, respectively}. }
\label{fig:overview}
\vspace{-3mm}
\end{figure*}

\subsection{DETR\nj{-}Based Highlight Detection Approaches}
DETR (Detection Transformer)~\cite{carion2020end} has provided a novel foundation for object detection by framing the task as a direct set prediction problem. Building upon this architecture, several models like Moment-DETR, QD-DETR, CG-DETR, and others~\cite{lei2021detecting, liu2022umt, lin2023univtg, moon2023query, moon2023correlation} have been proposed in the context of video highlight detection. These approaches leverage the self-attention mechanism inherent in the transformer architecture to process frame features, enabling the model to consider the entire sequence of video frames holistically.

In addition to self-attention across frame features, these models employ cross-modal attention to integrate the contextual information from natural language queries. The frame features are typically obtained by fusing the visual representations from a CLIP (ViT-B/32)~\cite{radford2021learning} encoder and temporal features from a SlowFast~\cite{feichtenhofer2019slowfast} model for each frame. The feature representation of the natural language queries is also extracted using the CLIP encoder, ensuring compatibility between the modalities.

These DETR-based approaches are trained with the dual objective of detecting video highlights and retrieving specific moments. By doing so, they are able to localize the most significant segments within a video stream that correspond to a given textual query, demonstrating advanced capabilities in both understanding and indexing video content.

\subsection{Efficient Training for Highlight Detection}
There exists a line of research to enhance the training efficiency of video highlight detection frameworks.
Among them, the prompt tuning methods~\cite{zhou2022learning, zhou2022conditional} were suggested as a way to efficiently exploit the pre-trained knowledge out of a model. In \nj{the} video highlight domain, the Visual Context Learner~(VCL)~\cite{vcl} was suggested to adapt \nj{the} prompt tuning strategy. The VCL train\nj{s} the trainable soft prompts prepended at the text input representation. The VCL is a detecter-free architecture that applied \nj{the} Context Optimization (CoOp)~\cite{zhou2022learning} framework while freezing \nj{the} whole pre-trained model and \nj{using a} simple network to localize the salient moments. This parameter-efficient approach showed improved performance while only using approximately \(2,000\) trainable parameters. This work showed \nj{the} possibility that CLIP itself can work as a highlight detector. From this line of work, we suggest HL-CLIP a finetuning strategy for CLIP on \nj{the} video highlight detection task that aims to maximize the potential of \nj{the} pre-trained model.

% \begin{figure*}[t!]
% \centering
% \includegraphics[width=.8\textwidth]{figures/arch_hr.png}
% \caption{The overall architecture of Highlight-CLIP~(HL-CLIP). Last few layers of transformer encoders are finetuned to estimate scaliency score between frame and query. \sh{green, blue가 gt인지 prediction인지 한문장.}}
% \label{fig:overview}
% \end{figure*}

\section{Method} \label{method}
\subsection{Motivation}

\sh{Since the pre-trained multimodal encoders contain \textit{implicitly}-trained temporal knowledge derived from the large-scale datasets, they achieve promising results in video highlight detection, which requires understanding of temporal information.
Based on this underlying assumption, we can expect that \textit{explicitly} providing temporal information by additional training will lead to \nj{both} better \nj{generalization} and \nj{better performance} in \nj{time-related} tasks.}
% Our approach of underlying assumption is pre-trained multimodal encoders would still be employed for video highlight detection tasks by \textit{implicit} temporal knowledge in pre-trained multimodal encoders, suggesting that employing \textit{explicitly} training temporal knowledge would be helpful for better generalization in temporal knowledge.
To this end, we introduce finetuning strategy to enhance temporal knowledge,
% which leverages temporal aspect, thereby
allowing us to enrich the specialized temporal knowledge for video highlight detection tasks.
\sh{The overall architecture is shown in \cref{fig:overview}.}

% Our approach is grounded in the premise that pre-trained multimodal encoders, despite not being explicitly trained on temporal data, can still be effectively harnessed for video highlight detection tasks. The challenge lies in adapting these models, which are rich in generalizable knowledge from vast and diverse datasets, to recognize and emphasize the temporal nuances essential for identifying highlights in video sequences. Acknowledging the difficulty of training generalizable knowledge with limited video-related datasets, we propose a finetuning strategy that leverages the latent spatial features learned by the encoder and introduces a temporal aspect suited for highlight detection.

\subsection{Task Definition}
The task of our study aims to identify the most salient and relevant moments aligned with a given user query \sh{in a video}, namely \nj{the} highlight detection task. The QVHighlight dataset \sh{consists} of \nj{the} diverse collection \nj{of} 10,000 videos with human-written free-form natural language queries~\cite{lei2021detecting}. \nj{Each query is} associated with specific moments of \nj{various} length, labeled with their relevance and saliency. Note that the main difference between moment retrieval and highlight detection is that highlight detection deals with identifying the specific moment related to \nj{a} query which is annotated \nj{with} high saliency by human annotators while moment retrieval finds all relevant segments in a video. 

Now, suppose that we have \nj{a set of videos $\mathcal{V}$, a set of queries $\mathcal{Q}$ and a set of saliency scores $\mathcal{Y}$ each consisting of $N$ elements as follows:} % corresponding to \nj{the} video.
% Each video is consisted of $k$ frames $x_1, \cdots, x_k \in \mathbb{R}^{K \times (w \times h \times 3)}$.
%\sh{Each video $V$ consists of $K$ frames, \ie $V \in \mathbb{R}^{K \times (w \times h \times 3)}$.}
\begin{comment}
    @오웬 Y에 대한 정의만 대충 적어주세용.
    쿼리 에 대한 건 추가 예정
\end{comment}
% The focus of our study is on the challenge of highlight detection within videos, a task that aims to identify the most salient and engaging moments that align with a specific user query. The QVHighlight dataset has diverse collection of over 10,000 videos, each accompanied by free-form, natural language queries. For each query, there are associated video moments of varying lengths, which are marked according to their relevance and saliency with respect to the query. Unlike moment retrieval, which seeks to find all relevant instances in a video, highlight detection focuses on segments that not only match the query but are also deemed highly salient by human annotators, indicating their significance or representativeness for the query. Video and query samples are defined as follows:
%
\begin{equation}
\begin{aligned}
\mathcal{V} &= \{V_i\}_{i=1}^{N}, \quad V_i = \{x_{i,j}\}_{j=1}^{K}, \\
\mathcal{Q} &= \{q_i\}_{i=1}^{N},\\
\mathcal{Y} &= \{Y_i\}_{i=1}^{N}, \quad Y_i = \{y_{i,j}\}_{j=1}^{K}.
\end{aligned}
\end{equation}
\nj{Here, \( \mathcal{V} \) contains \( N \) videos, each represented by \( V_i \), which consists of \( K \) frames \( x_{i,j} \).} The index \( i \) ranges from 1 to \( N \), identifying each video in the dataset, and \( j \) ranges from 1 to \( K \), indicating the frame \sh{index} within a video. Additionally, \( \mathcal{Q} \) is the set of query sentences associated with the videos, with \( q_i \) being the query corresponding to \sh{$V_i$}. \( \mathcal{Y} \) denotes the set of saliency scores for \(N\) videos, each represented by \(Y_i\) is the saliency scores of each video labeled by human annotators. \(y_{i,j}\) is \sh{the ground-truth} saliency score of \nj{the} \(j\)-th frame of \nj{the} \(i\)-th video. The saliency score is originally annotated in a five-point Likert-scale. We normalized this scale to \(0\) to \(1\). However, not all frames have a saliency score, and only salient moments are labeled. For training, we set the saliency score of unlabeled frames to `Very Bad'. 

% \subsection{Proposed Method}
\subsection{\sh{CLIP as a Highlight Detector}} \label{hl_clip}
Our approach to highlight detection is unique in that it operates without a dedicated highlight detector module. Instead, we \sh{fine}tuned \nj{a} multimodal encoder to process both visual and textual input, thus allowing the model to infer the saliency of different video segments directly from the query and the video content itself. Through this lens, we aim to advance the field of video analysis by developing a model that can intuitively and effectively pinpoint highlights in a vast array of video content.

Unlike previous methods using both CLIP and SlowFast~\cite{feichtenhofer2019slowfast} features to train separate highlight detector\nj{s}, we use \sh{only} CLIP (ViT-B/32)'s visual encoder and text encoder to \sh{perform} the highlight detection \nj{task} \sh{as follows:}
% Through this approach, we have achieved the best highlight detection score to the best of our knowledge.
% However, the challenge remains in that our approach cannot localize the moment with a given query in the start and end time due to its structural limitation.
% We unfreezed the last two layers of CLIP (ViT-B/32) to \sh{perform} the highlight detection as following:
%
\begin{equation}
\begin{aligned} 
h^{v}_{i,j} &= \mathcal{F}_\text{vision}(x_{i,j}; \theta_\text{vision}) \\
h^{t}_i &= \mathcal{F}_\text{text}(q_i; \theta_\text{text}),
\end{aligned}
\end{equation}
where \( h^{v}_{i,j} \) denotes the visual feature vector for the \( j \)-th frame of the \( i \)-th video, $\mathcal{F}_\text{vision}(\cdot)$ is the visual encoder parameterized by $\theta_\text{vision}$, \( x_{i,j} \) represents the \( j \)-th frame of the \( i \)-th video, \( h^{t}_i \) denotes the text feature vector for the \( i \)-th video, $\mathcal{F}_\text{text}(\cdot)$ is the text encoder, and \( q_i \) is the query sentence associated with the \( i \)-th video.
As shown in \cref{fig:overview}, we finetuned the last two layers of both encoders.

Then the cosine similarity \sh{between} \( h^{v}_{i,j} \) and \( h^{t}_i \)
% , \(\hat{y}_{i,j}\) 
is used as \nj{the} predicted saliency score for estimating relevancy between the video frame and \nj{the} given query\sh{:}
\begin{equation}
\hat{y}_{i,j} = \text{sim}(h^{v}_{i,j}, h^{t}_i),
\end{equation}
where \sh{the cosine similarity }\(\hat{y}_{i,j}\) \sh{indicates} the saliency of \(x_{i,j}\) given query \(q_i\).
\sh{Note that $h^{t}_i$ is broadcast across the $K$ frames of $V_i$.}

\sh{However, although we achieve \nj{a} highly competitive performance in highlight detection, the challenge remains in that our approach has difficulty in moment retrieval due to its structural limitation in localizing the moment with a given query in the start and end time.}

\subsection{Training \sh{and Inference}} \label{train_scheme}
Specifically, our approach involves finetuning the last few layers of a pre-trained CLIP model with \nj{a} video highlight dataset.
% We introduce a process of average pooling over neighboring frame features to capture the temporal continuity inherent in video streams.
Given a video sample consisting of \sh{$K$} frames, we observe that adjacent frames exhibit high similarity due to their sequential nature. To capitalize on this characteristic, we arrange the frame features in a batch-wise stack of dimensions \(\sh{K} \times N\), where \(N\) is the number of video samples. %and \(75\times N\) is the batch size. 
This configuration aids in distinguishing the subtle differences between similar-looking frames, effectively allowing the model to discern the salient moments that constitute a video highlight.

Additionally, to ensure alignment between the video frames and the textual queries, we replicate the query feature vector \sh{$K$} times to match the temporal dimension of the video sample, resulting in a similar \(\sh{K} \times N\) structure. This stacking strategy enables the model to learn in a manner that sharpens its focus on the informative terms within the query that are most relevant for highlight detection.

% \subsection{Loss Function}
\paragraph{Loss function.}
To train \sh{our HL-CLIP}, we used mean squared error (MSE) between predicted saliency score and ground truth \sh{as follows:}

\begin{equation}
\mathcal{L}_{\text{saliency}} = \frac{1}{N \cdot K} \sum_{i=1}^{N} \sum_{j=1}^{K} (\hat{y}_{i,j} - y_{i,j})^2.
\end{equation}

% where \(\hat{y}_{i,j}\) denotes predicted saliency score of \(j\)-th frame of \(i\)-th video and \(y_{i,j}\) denotes the ground truth saliency score.
% 얘는 ln127이랑 ln91에서 이미 말해줘서 안 써도 될 것 같습니다.

\sh{
\paragraph{Inference.}
At the inference stage, we use the temporal-aggregated saliency score derived by the \textbf{saliency pooling}, \ie simply average-pooling a set of saliency scores from neighboring frames.
By the simple pooling technique, we consider the semantic similarities among the adjacent frames and estimate the saliency score more robustly.
% Kindly refer to the related experiment in \cref{fintune}.
}

\begingroup
\setlength{\tabcolsep}{3.9pt} % Default value: 6pt
\renewcommand{\arraystretch}{0.95} % Default value: 1
\begin{table*}[t!]
	\centering
	{\small 
 \resizebox{0.9\linewidth}{!}{
        \begin{tabular}{l|ccccccc|ccccccc}
        \Xhline{4\arrayrulewidth}
\multicolumn{1}{c|}{Split} & \multicolumn{7}{c|}{test} & \multicolumn{7}{c}{val} \\ \hline
\multicolumn{1}{c|}{\multirow{3}{*}{Method}} &  \multicolumn{5}{c}{MR}     & \multicolumn{2}{c|}{HD} & \multicolumn{5}{c}{MR}     & \multicolumn{2}{c}{HD} \\ \cline{2-15} 
\multicolumn{1}{c|}{} & \multicolumn{2}{c}{R1} & \multicolumn{3}{c}{mAP}  & \multicolumn{2}{c|}{\textgreater{}= Very Good} & \multicolumn{2}{c}{R1} & \multicolumn{3}{c}{mAP}  & \multicolumn{2}{c}{\textgreater{}= Very Good} \\ \cline{2-15} 
\multicolumn{1}{c|}{} & @0.5 & @0.7 & @0.5 & @0.75 & Avg. & mAP & HIT@1 & @0.5 & @0.7 & @0.5 & @0.75 & Avg. & mAP & HIT@1\\ \Xhline{4\arrayrulewidth}
BeautyThumb~\cite{song2016click}  &  -& -& -& -& -& 14.36 & 20.88 & - & - & - & - & - & - & -\\
DVSE~\cite{liu2015multi}& -& -& -& -& -& 18.75 & 21.79 & - & - & - & - & - & - & -\\
MCN~\cite{anne2017localizing} & 11.41& 2.72 & 24.94& 8.22 & 10.67& - & - & - & - & - & - & - & - & -\\
CAL~\cite{escorcia2019temporal} & 25.49& 11.54& 23.40& 7.65 & 9.89 & - & - & - & - & - & - & - & - & -\\
XML~\cite{lei2020tvr} & 41.83& 30.35& 44.63& 31.73& 32.14& 34.49 & 55.25 & - & - & - & - & - & - & -\\
XML+\cite{lei2020tvr} & 46.69& 33.46& 47.89& 34.67& 34.90& 35.38 & 55.06 & - & - & - & - & - & - & -\\ \hline
% \multicolumn{9}{c}{DETR decoder} \\ \cline{1-9} 
Moment-DETR~\cite{lei2021detecting}& 52.89 & 33.02 & 54.82 & 29.40 & 30.73 & 35.69 & 55.60 & 53.94 & 34.84 & - & - & 32.20 & 35.65 & 55.55 \\
UMT~\cite{umt} & 56.23& 41.18& 53.38& 37.01& 36.12& 38.18 & 59.99 & 60.26 & 44.26 & - & - & 38.59 & 39.85 & 64.19 \\
QD-DETR~\cite{moon2023query} & 62.40& 44.98 & 62.52 & 39.88 & 39.86 & 38.94 & 62.40 & 62.68 & 46.66 & 62.23 & 41.82 & 41.22 & 39.13 & 63.03 \\ 
UniVGT~\cite{univtg} & 58.86 & 40.86 & 57.6 & 35.59 & 35.47 & 38.20 & 60.96&  59.74 & - & - & - & 36.13 & 38.83 & 61.81 \\ 
EaTR~\cite{eatr} & - & - & - & - & - & - & - & 61.36 & 45.79 & 61.86 & 41.91 & 41.74 & 37.15 & 58.65 \\
VCL~\cite{vcl} & - & - & - & - & - & - & - & 43.33 & 25.75 & 39.23 & 24.95 & 21.13 & 38.84 & 65.74\\
CG-DETR~\cite{moon2023correlation} & \textbf{65.43} & \textbf{48.38} & \textbf{64.51} & \textbf{42.77} & \textbf{42.86} & 40.33 & 66.21& \textbf{67.35} & \textbf{52.06} & \textbf{65.57} & \textbf{45.73} & \textbf{44.93} & 40.79 & 66.71 \\ 
\hline
HL-CLIP~(Ours) & - & - & - & - & - & \textbf{41.94} & \textbf{70.60} & - & - & - & - & - & \textbf{42.37} & \textbf{72.40} \\
\Xhline{4\arrayrulewidth}
\end{tabular}
}
\vspace{-2mm}
	\caption{
 % Performance comparison on QVHighlights. Avg. mAP is calculated with IoU thresholds ranging from 0.5 to 0.95 in 0.05 intervals.
 Performance comparison on QVHighlights \textit{test} and \textit{val} splits. We report the average highlight detection score of 3 runs with different seeds on validation and test split. We evaluated with HL-CLIP-2 (SP) our best configuration referred on \cref{fintune}. \sh{Although our }HL-CLIP is not capable of \sh{performing} moment retrieval as mentioned in \cref{hl_clip}, \sh{we conjecture that our HL-CLIP has much potential in the moment retrieval task as well considering its superior performance in highlight detection}. \sh{We \nj{leave} its application to the moment retrieval as a future work.}
 % Moment retrieval is left for future work.
 }
	\label{table_QVHighlight}
}
\vspace{-3mm}
\end{table*}
\endgroup

% run1: test_HL-min-VeryGood-mAP: 41.94 / test_HL-min-VeryGood-Hit1: 70.62
%       val_HL-min-VeryGood-mAP: 42.32 / val_HL-min-VeryGood-Hit1: 72.19
% run2: test_HL-min-VeryGood-mAP: 41.97 / test_HL-min-VeryGood-Hit1: 70.69
%       val_HL-min-VeryGood-mAP: 42.42 / val_HL-min-VeryGood-Hit1: 72.58
% run3: test_HL-min-VeryGood-mAP: 41.93 / test_HL-min-VeryGood-Hit1: 70.49
%       val_HL-min-VeryGood-mAP: 42.39 / val_HL-min-VeryGood-Hit1: 72.45

% \subsection{Saliency Pooling}
% At the inference stage, we average pooled the saliency score with the neighboring frames. This turns out to boost the performance by quite a margin. We also tried to apply the pooling in various stages to distill the temporal information to the multimodal visual encoder. We tried to average pool the temporally neighboring input pixel frames and also tried to average pool the output feature of the multimodal visual encoder while training. However, applying the average pooling showed the most significant performance boost in both the validation and test split of QVHighlight. The performance boost effect of pooling is referred to in Section~\ref{fintune}.

\section{Experiment}

The experimental section of our work presents an initial exploration into the adaptation of pre-trained multimodal encoders for the task of video highlight detection.
Given the brevity of this short paper, our focus is to provide a snapshot of ongoing research efforts rather than an exhaustive evaluation.
Each subsection offers insights into different facets of our approach and reflects the potential of these methods to enhance video analysis tasks.

\begin{table}[t!]
\begin{center}
\resizebox{0.8\linewidth}{!}{
\begin{tabular}{c|c|c}
\Xhline{4\arrayrulewidth}
Method & mAP (Std.) & HIT@1 (Std.) \\
\hline
CLIP & 37.01  & 63.03 \\
HL-CLIP-1 & 39.70 (0.03) & 67.69 (0.03) \\ % 39.7/67.68,  39.65/68.19, 39.73/67.48, 39.74/67.42
HL-CLIP-1 (SP) & 41.97 (0.2) & 71.64 (0.1) \\ % 42.12/71.55, 41.98/71.81, 41.63/71.68, 42.15/71.55
HL-CLIP-2 & 40.00 (0.06) & 68.99 (0.28) \\ % 40.01/68.84, 39.91/68.9, 40.0/68.77, 40.09/69.48
HL-CLIP-2 (SP) & \bf 42.40 (0.05) & \bf 72.42 (0.25)  \\ % 42.45/72.39, 42.4/72.84, 42.45/72.26, 42.32/72.19
\Xhline{4\arrayrulewidth}
\end{tabular}
}
\end{center}
\vspace{-4mm}
\caption{\textbf{Comparison with multiple finetuning variants.} Highlight Detection ($>=$Very Good) on QVHighlights \textit{val} split. We reported standard deviation of 4 runs with different seeds. SP denotes for Saliency Pooling.}
\label{tab:ft}
\vspace{-3mm}
\end{table}

\subsection{CLIP Finetuning} \label{fintune}
% As mentioned in Section \ref{method}, we \sh{finetuned} part of the CLIP encoder \sh{with} the video highlight dataset. We sampled frame per two seconds to extract the feature. There are two variants we tried to finetune. First, we finetuned only the single last layer of the transformer encoder at CLIP (ViT-B/32) which is HL-CLIP-1. Second, we finetuned the last two unfreezed layers of CLIP, \sh{\ie HL-CLIP-2}. In \cref{tab:ft}, HL-CLIP-2 showed better performance than HL-CLIP-1. And applying saliency pooling (HL-CLIP-2-SP) achieved the best performance in highlight detection task. Our HL-CLIP achieves good performance but is detector-free, using fewer parameters to forward. 

In \cref{tab:ft}, the quantitative results demonstrate that HL-CLIP-2 outperforms HL-CLIP-1 with a mean Average Precision (mAP) of 40.00 compared to 39.70, and a HIT@1 score of 68.99 and 67.69, respectively. Notably, the integration of saliency pooling further enhances performance, as seen with HL-CLIP-2-SP achieving the highest mAP of 42.40 and HIT@1 of 72.42, confirming its superior efficacy in the highlight detection task. Our HL-CLIP achieves \sh{promising results within a detector-free framework,} using fewer parameters. 

\subsection{Comparison with Baselines} \label{comparison}
In \cref{table_QVHighlight}, HL-CLIP achieved the state-of-the-art performance among the previously suggested methods on highlight detection task for both \textit{test} and \textit{val} splits.
Due to \sh{the} structural characteristic, \sh{our }HL-CLIP cannot directly localize the specific moment with a given queries~(Moment Retrieval task).
However, \sh{since the }moment retrieval \sh{task} is highly relevant to selecting the most salient moment, \ie highlight detection, \sh{there exists a} possibility of adapting HL-CLIP to the moment retrieval task with further refinement and auxiliary mechanisms.

\section{Conclusion}  
In this work, we explore the potential of CLIP as a video highlight detector.
% Thanks to its inherent knowledge for temporal information, our proposed HL-CLIP is equipped with outstanding ability for highlight detection by simply finetuning part of the CLIP and saliency pooling technique.
\sh{Thanks to its general knowledge from large-scale dataset, our proposed HL-CLIP is equipped with outstanding ability for highlight detection by simply finetuning part of the CLIP and saliency pooling technique.}
We achieve the state-of-the-art performance on the representative benchmark QVHighlight, showing the competitiveness of CLIP-only framework.

\paragraph{Future work.}
\sh{Our current framework can be extended} to advance the model's capabilities in complex video understanding tasks, \eg leveraging the finetuned encoder to target precise moment retrieval. Based on our work, we plan to suggest score-based efficient network that can localize the moment with a saliency score and finetuned features.

{
    \small
    \bibliographystyle{ieeenat_fullname}
    \bibliography{main}
}


\end{document}